\documentclass[11pt]{article}

\usepackage[final]{acl}

\usepackage{times}
\usepackage{latexsym}

\usepackage[T1]{fontenc}

\usepackage[utf8]{inputenc}

\usepackage{microtype}

\usepackage{inconsolata}

\usepackage{graphicx}

\usepackage{booktabs}     
\usepackage{array}
\usepackage{tabularx}
\usepackage{ragged2e}   
\usepackage{makecell}   
\usepackage{adjustbox}
\usepackage{pifont} 
\usepackage[table]{xcolor} 
\usepackage{amsmath}
\usepackage{marvosym}

\title{AutoFigure-Edit: Generating Editable Scientific Illustration}

\author{
 \textbf{Zhen Lin\textsuperscript{1*}},
 \textbf{Qiujie Xie\textsuperscript{2,1*}},
 \textbf{Minjun Zhu\textsuperscript{2,1*}},
 \textbf{Shichen Li\textsuperscript{1,3}},
 \textbf{Qiyao Sun\textsuperscript{1}},
\\
 \textbf{Enhao Gu\textsuperscript{1}},
 \textbf{Yiran Ding\textsuperscript{1}},
 \textbf{Ke Sun\textsuperscript{1}},
 \textbf{Fang Guo\textsuperscript{1}},
 \textbf{Panzhong Lu\textsuperscript{1}},
\\
 \textbf{Zhiyuan Ning\textsuperscript{1}},
 \textbf{Yixuan Weng\textsuperscript{1}$^{\dagger}$},
 \textbf{Yue Zhang\textsuperscript{1}\textsuperscript{\Letter}}
\\
\\
 \textsuperscript{1}Engineering School, Westlake University,
 \textsuperscript{2}Zhejiang University,
 \textsuperscript{3}Soochow University
\\
 \small{
 \textbf{*: Equal contribution $\dagger$: Project leader.}
 }
\\
\small{
 \textbf{Correspondence author\Letter:} \href{zhangyue@westlake.edu.cn}{zhangyue@westlake.edu.cn}
 }
}

\begin{document}
\maketitle
\begin{abstract}

High-quality scientific illustrations are essential for communicating complex scientific and technical concepts, yet existing automated systems remain limited in editability, stylistic controllability, and efficiency. We present \textsc{AutoFigure-Edit}, an end-to-end system that generates fully editable scientific illustrations from long-form scientific text while enabling flexible style adaptation through user-provided reference images. By combining long-context understanding, reference-guided styling, and native SVG editing, it enables efficient creation and refinement of high-quality scientific illustrations. To facilitate further progress in this field, we release the video\footnote{\url{https://youtu.be/10IH8SyJjAQ}}, full codebase\footnote{\url{https://github.com/ResearAI/AutoFigure-Edit}} and provide a website for easy access and interactive use\footnote{\url{https://deepscientist.cc/}}.

\end{abstract}

\section{Introduction}

Creating high-quality scientific illustrations often takes researchers several days and demands both substantial domain expertise and professional-level design skills~\citep{huang2026scifigautomatingscientificfigure}. It requires a rigorous, logic-aware understanding of \textbf{long-form scientific texts} ($>$10k tokens), while the visual rendering must carefully balance \textbf{structural fidelity} and \textbf{image quality} to produce figures that are clear, accurate, and aesthetically appealing~\citep{chang2025sridbenchbenchmarkscientificresearch, zhu2026autofigure}.

Research on \textbf{automatically generating scientific illustrations} from long-form scientific texts remains limited. Code-as-intermediate approaches~\citep{belouadi2024automatikz, belouadi2024detikzify, belouadi2025tikzero, ellis2018learning} achieve strong geometric correctness, but often sacrifice visual aesthetics and readability~\citep{zhu2026autofigure}. Meanwhile, end-to-end mainstream Text-to-Image (T2I) models can produce visually appealing illustrations yet frequently fail to maintain \textbf{structural fidelity} on long scientific inputs~\citep{liu2025improving, huang2026longtexttoimage}. Therefore, directly transforming long scientific texts into illustrations that are both \textbf{accurate} and \textbf{visually compelling} remains challenging.

In our previous work, we introduced \textsc{AutoFigure}~\citep{zhu2026autofigure}, an agentic framework grounded in the Reasoned Rendering paradigm that produces accurate and visually appealing illustrations through an iterative refinement process. Despite its ability to automatically generate high-quality illustrations, \textsc{AutoFigure} has several limitations: (i) the generated visual elements are \textbf{fixed and non-editable}. Refinement can only be performed by modifying the user-provided textual prompt; (ii) generating illustrations in a desired style relies heavily on \textbf{prompt engineering}, which can be ambiguous and may lead to inaccurate or unintended stylistic outcomes; and (iii) its iterative sketch-to-render refinement \textbf{tightly couples layout planning with final raster rendering}, without exposing an explicit structural scaffold. As a result, fine-grained edits (e.g., layout adjustments) are difficult and text rendering is often unstable.

\begin{table*}[htbp]
\newcolumntype{Y}{>{\RaggedRight\arraybackslash}X}
\newcommand{\cmark}{\ding{51}} 
\newcommand{\xmark}{\ding{55}} 
\setlength{\tabcolsep}{4pt} 
\small
\centering
\begin{tabularx}{\textwidth}{@{} Y c c c c @{}}
\toprule
Method & Sci. Gen. & Editable & Style Control & GUI Support \\
\midrule
StarVector \& OmniSVG \citep{rodriguez2025starvector, yang2025omnisvg} & Limited & \cmark & Prompt (Hard) & Web Only \\
AutomaTikZ \citep{belouadi2024automatikz}                                    & \cmark  & \cmark & Prompt (Hard) & Web Only \\
DeTikZify \citep{belouadi2024detikzify}                                      & \cmark  & \cmark & Sketch (Hard) & Web Only \\
GPT-Image \citep{hurst2024gpt}                                               & \cmark  & \xmark & Prompt (Hard) & Web Only \\
Diagram Agent \citep{wei2025words}                                           & \cmark  & \cmark & Prompt (Hard) & None \\
PaperBanana \citep{zhu2026paperbananaautomatingacademicillustration}         & \cmark  & \xmark & Prompt (Hard) & Web Only \\
EditBanana \citep{editbanana}                                                & \xmark  & \cmark & \xmark        & Web Only \\
SciFig \& SciSketch \citep{huang2026scifigautomatingscientificfigure, wang-etal-2025-scisketch} & \cmark  & \cmark & Prompt (Hard) & None \\
VisPainter \citep{sun2025pixelspathsmultiagentframework}                     & \cmark  & \cmark & Prompt (Hard) & None \\
\textsc{AutoFigure} \citep{zhu2026autofigure}                                         & \cmark  & \xmark & Prompt (Hard) & Web Only \\
\midrule
\textbf{\textsc{AutoFigure-Edit} (Ours)}                                              & \textbf{\cmark} & \textbf{\cmark} & \textbf{Reference (Easy)} & \textbf{Web + Editor} \\
\bottomrule
\end{tabularx}
\vspace{-5pt}
\caption{Comparison of \textsc{AutoFigure-Edit} with other relevant systems. ``Sci. Gen.'' denotes scientific illustration generation; ``GUI Support'' denotes the level of integrated graphical user interface and editing capabilities.}
\label{tab:compare}
\vspace{-10pt}
\end{table*}

To address these limitations, we present a substantially enhanced system, named \textsc{AutoFigure-Edit}, that transforms long-form scientific text and a reference style image into a \textbf{fully editable} SVG illustration. It enables \textbf{reference-guided style control}, reducing reliance on ambiguous prompt engineering, and decouples layout planning from final rendering via an \textbf{explicit structural scaffold}, allowing layout edits directly on the vector scaffold without retrying the full sketch-to-render loop. Building on this design, \textsc{AutoFigure-Edit} provides the following features:

\textbf{Scientific Illustration Generation.} Directly transforms long-form scientific text into accurate, publication-quality illustrations. 

\textbf{Reference-Guided Style Control.} Enables controllable visual adaptation via a user-provided exemplar while preserving semantic structure. 

\textbf{Editable SVG with Embedded Visual Editor.} Produces structurally organized, component-level editable SVGs and supports real-time refinement through an integrated interactive canvas.

Quantitative experiments and user studies (Section \ref{sec:evaluation}) demonstrate the effectiveness and practical value of \textsc{AutoFigure-Edit} in generating high-quality, editable scientific illustrations.

\section{Related Work}

\textbf{Automation of scientific illustration} has evolved from simple summarization to complex synthesis. However, achieving a balance between generation quality and post-generation editability remains challenging. Existing text-to-figure systems~\citep{zhu2026paperbananaautomatingacademicillustration,huang2026scifigautomatingscientificfigure,zhu2026autofigure} can automatically create high-quality illustrations from long textual descriptions, but they typically produce static outputs with limited support for iterative refinement, requiring full regeneration for even minor adjustments. To improve customizability, recent approaches convert rasterized renderings into vector representations. Nevertheless, their reliance on pre-rendered pixel inputs can lead to semantic information loss \citet{sun2025pixelspathsmultiagentframework}. Meanwhile, editing tools such as EditBanana \citep{editbanana} provide post-hoc modification capabilities, reling on externally provided images as inputs. In contrast, \textbf{\textsc{AutoFigure-Edit} offers a unified, end-to-end pipeline} that not only generates illustrations from scratch but also represents all components as fully editable objects, enabling precise control.

\textbf{Programmatic Synthesis.}
Despite the proficiency of diffusion models \citep{saharia2022photorealistic} in general visual synthesis, their limited structural transparency make them ill-suited to the strict compositional constraints of scientific figures. To enhance controllability, recent work has explored Text-to-Code-to-Image pipelines that use programmatic representations (e.g., TikZ or SVG) as intermediate forms \citep{belouadi2024automatikz, belouadi2024detikzify}. However, purely programmatic generation is often brittle. Small syntax errors can trigger rendering failures, and the absence of an intuitive visual editing interface increases the effort required for iterative refinement.
\textsc{AutoFigure-Edit} addresses these issues by combining long-form context understanding with robust structural reconstruction, striking better balances between stylistic flexibility and editability. The comparison with relevant systems is shown in Table~\ref{tab:compare}.

\section{AutoFigure-Edit}
\begin{figure*}[t]
    \centering
    \includegraphics[width=1.0\linewidth]{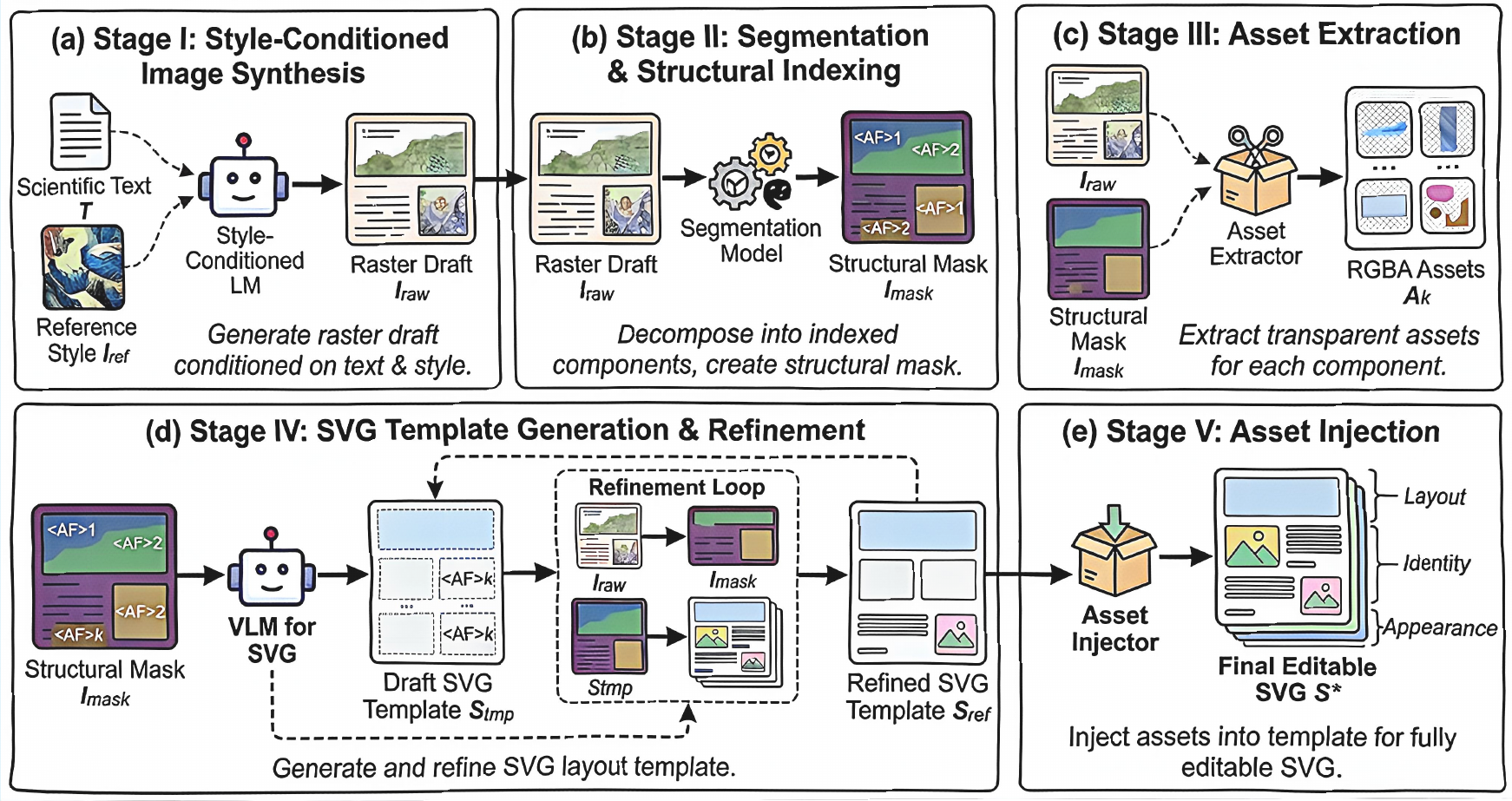}
    \caption{An overview of \textsc{AutoFigure-Edit}. This figure is also produced by \textsc{AutoFigure-Edit} and serves as a qualitative showcase of its generation quality.}
    \label{fig:method}
    \vspace{-10pt}
\end{figure*}

We introduce \textsc{AutoFigure-Edit} (Figure~\ref{fig:method}), an automated system that transforms long-form scientific text into structured, fully editable scientific illustrations and supports flexible style adaptation via user-provided reference images. 

\subsection{Framework Overview}

The task of automated scientific illustration generation involves reconciling three competing goals: (i) \textbf{semantic faithfulness} to the text; (ii) \textbf{stylistic consistency} with the reference image; (iii) explicit structural decomposition to support \textbf{downstream editing}. Formally, given a long-form scientific text $T$ and a reference style image $I^{\text{ref}}$, the objective is to learn a mapping: 
\[
S^{\star} = \mathcal{F}(T, I^{\text{ref}}),
\]
where $S^{\star}$ is an editable vector graphic that preserves the semantics of $T$ while conforming to the visual style of $I^{\text{ref}}$.

Directly parameterizing this mapping is ill-posed due to the absence of explicit structural supervision and the entanglement of layout, instance, and visual appearance. We therefore \textbf{decompose the transformation into sequential stages} that progressively derive structure from an intermediate raster draft, disentangling layout planning, object identity, and visual rendering while preserving semantic and stylistic coherence.

\subsubsection{Stage I: Style-Conditioned Image Synthesis}

We first generate a raster draft $I^{\text{raw}}$ conditioned jointly on the input scientific text and a reference style image, using a style-conditioned text-to-image model (e.g., Gemini-3-Pro-Image-Preview). This stage translates textual descriptions into explicit visual entities while incorporating high-level stylistic cues from the reference image, thereby establishing semantic–stylistic alignment.

\subsubsection{Stage II: Segmentation and Structural Indexing}

To expose the compositional structure of the raster draft, we then apply instance segmentation~\citep{carion2025sam} to \textbf{decompose $I^{\text{raw}}$ into a set of visual components} $\{M_k, B_k\}_{k=1}^{K}$, where $M_k$ and $B_k$ denote the mask and bounding box of the $k$-th component, respectively. 
Instead of retaining the original appearance of each region, we construct a simplified structural rendering in which every instance is filled with a uniform tone and assigned a unique identifier token (e.g., \texttt{<AF>k}). By suppressing texture and color while preserving spatial configuration and instance identity, this transformation converts the raster image into \textbf{an indexed structural layout}, providing a coordinate-aware scaffold for subsequent vector generation.

\subsubsection{Stage III: Asset Extraction}

While Stage II enables explicit structural information, appearance cues must be preserved to ensure faithful reconstruction. Therefore, for each segmented instance $(I^{\text{raw}}, B_k)$, we extract the corresponding visual content and remove the background~\citep{BiRefNet} to obtain a transparent RGBA asset $A_k$. This process decouples geometric placement from visual texture, storing appearance as standalone icon-like assets while delegating spatial organization to the structural scaffold. As a result, subsequent layout modifications can be performed without altering stylistic details.

\begin{figure*}[ht]
    \centering
    \includegraphics[width=\linewidth]{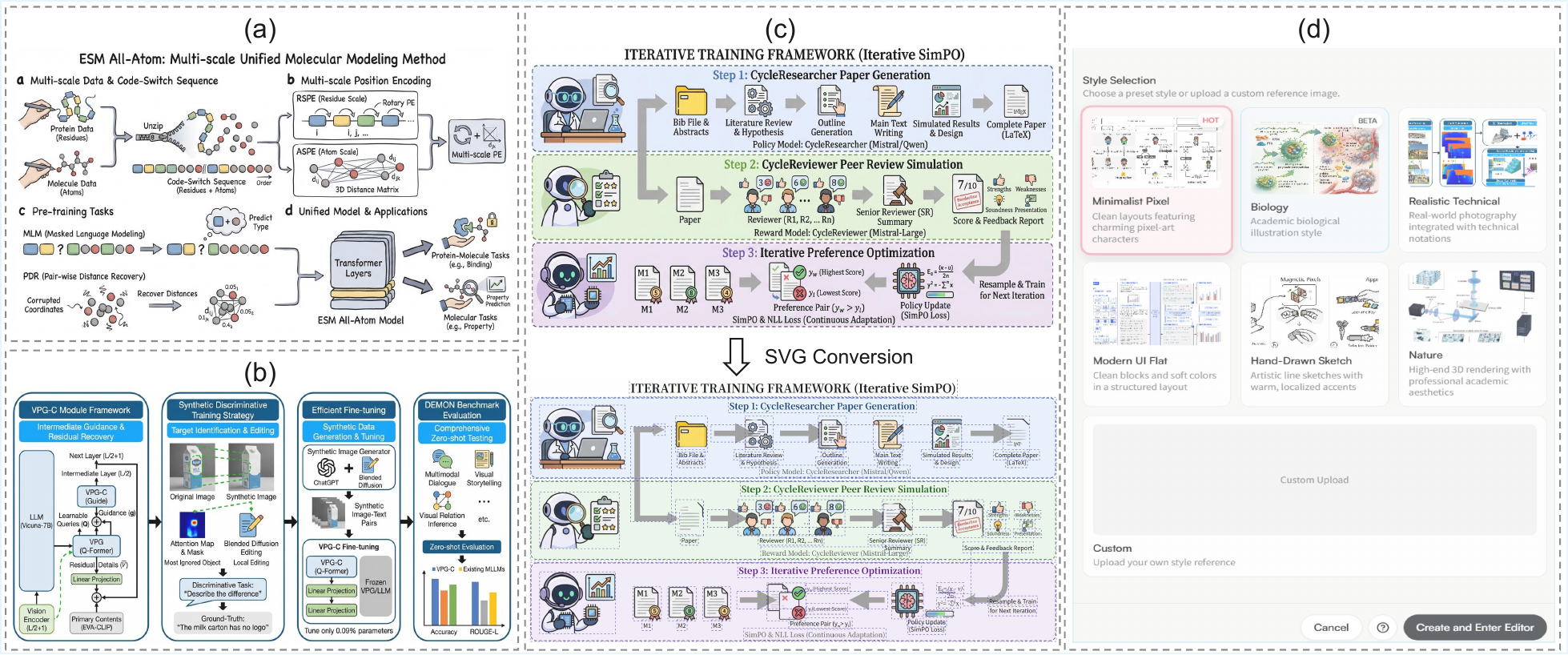}
    \caption{Representative outputs of \textsc{AutoFigure-Edit}. (a)-(b) are bitmap (PNG) figures generated from long-form scientific descriptions across various domains. (c) shows the PNG-to-SVG conversion case of \textsc{AutoFigure-Edit}, including the original bitmap (bottom) and its corresponding vectorized SVG result (down). (d) is the web interface of \textsc{AutoFigure-Edit}, allowing users to select predefined style templates or upload custom reference images.
    }
    \label{fig:edit_case}
    \vspace{-10pt}
\end{figure*}

\subsubsection{Stage IV: SVG Template Generation and Refinement}
Given the indexed structural representation $I^{\text{mask}}$, we then prompt a vision-language model (e.g., Gemini-3.1-Pro-Preview) to generate an \textbf{SVG layout template} $S^{\text{tmp}}$ containing placeholder elements aligned with the \texttt{<AF>k} identifiers. 

To improve alignment with the target figure, we further perform a \textbf{lightweight refinement step} by re-prompting the vision-language model with the original raster draft, the structural mask, a rendered preview of the current SVG, and the corresponding SVG code. The model is instructed to correct discrepancies in two aspects: \textbf{positional consistency} (icon placement, text alignment, arrows, lines) and \textbf{stylistic consistency} (proportions, fonts, stroke widths, and colors). Identifier mappings and placeholder group structures are preserved to ensure compatibility with subsequent asset injection. In practice, the refinement process requires only 0–2 iterations to obtain a satisfactory template $S^{\text{ref}}$.

\subsubsection{Stage V: Asset Injection}

Finally, the extracted appearance assets $\{A_k\}_{k=1}^{K}$ are injected into the refined SVG template $S^{\text{ref}}$ by replacing each placeholder with its corresponding asset. This process produces \textbf{a fully editable SVG $S^{\star}$} where layout, object identity, and visual appearance \textbf{remain independently manipulable.} Users can subsequently modify geometry, adjust style, or update individual components without disrupting the overall composition.

\subsection{Applications}

We demonstrate the utility of \textsc{AutoFigure-Edit} through three representative application scenarios, showcasing its generation capability, stylistic adaptability, and editability. Beyond being a technical system, \textsc{AutoFigure-Edit} serves as a \textbf{productivity tool for researchers} across fields, lowering the barrier to high-quality scientific illustrations.

\textbf{High-quality Scientific Illustration Generation.}
The primary use case of \textsc{AutoFigure-Edit} is the generation of publication-level illustrations directly from long-form scientific texts. Given a method section or system overview spanning thousands of tokens, the system automatically extracts the key entities, relations, and procedural stages, and transforms them into illustrations that are both \textbf{accurate} and \textbf{visually appealing.} This capability substantially \textbf{reduces the time and expertise} traditionally required to translate dense technical content into clear visual form. In Figure~\ref{fig:edit_case} (a)-(c), we present representative examples of the generated results, demonstrating both semantic fidelity to the source text and high visual quality.

\textbf{Style Adaptation.}

Given a user-provided reference image, \textsc{AutoFigure-Edit} can adapt to \textbf{a wide range of visual styles}, changing color palettes, typography, icon aesthetics, spacing density, and visual hierarchy while \textbf{preserving semantic structure.} Rather than relying on prompt-level style descriptions, the system explicitly conditions on a visual exemplar and transfers high-level stylistic attributes in a controlled manner. This enables users to experiment with multiple visual appearances of the same scientific content, facilitating alignment with venue- or lab-specific visual standards and reducing reliance on manual graphic design expertise. In Figure~\ref{fig:edit_case} (d), we illustrate the input interface of \textsc{AutoFigure-Edit}, where users provide both the source text and a reference style image to guide generation.

\begin{figure}[t]
    \centering
    \includegraphics[width=\linewidth]{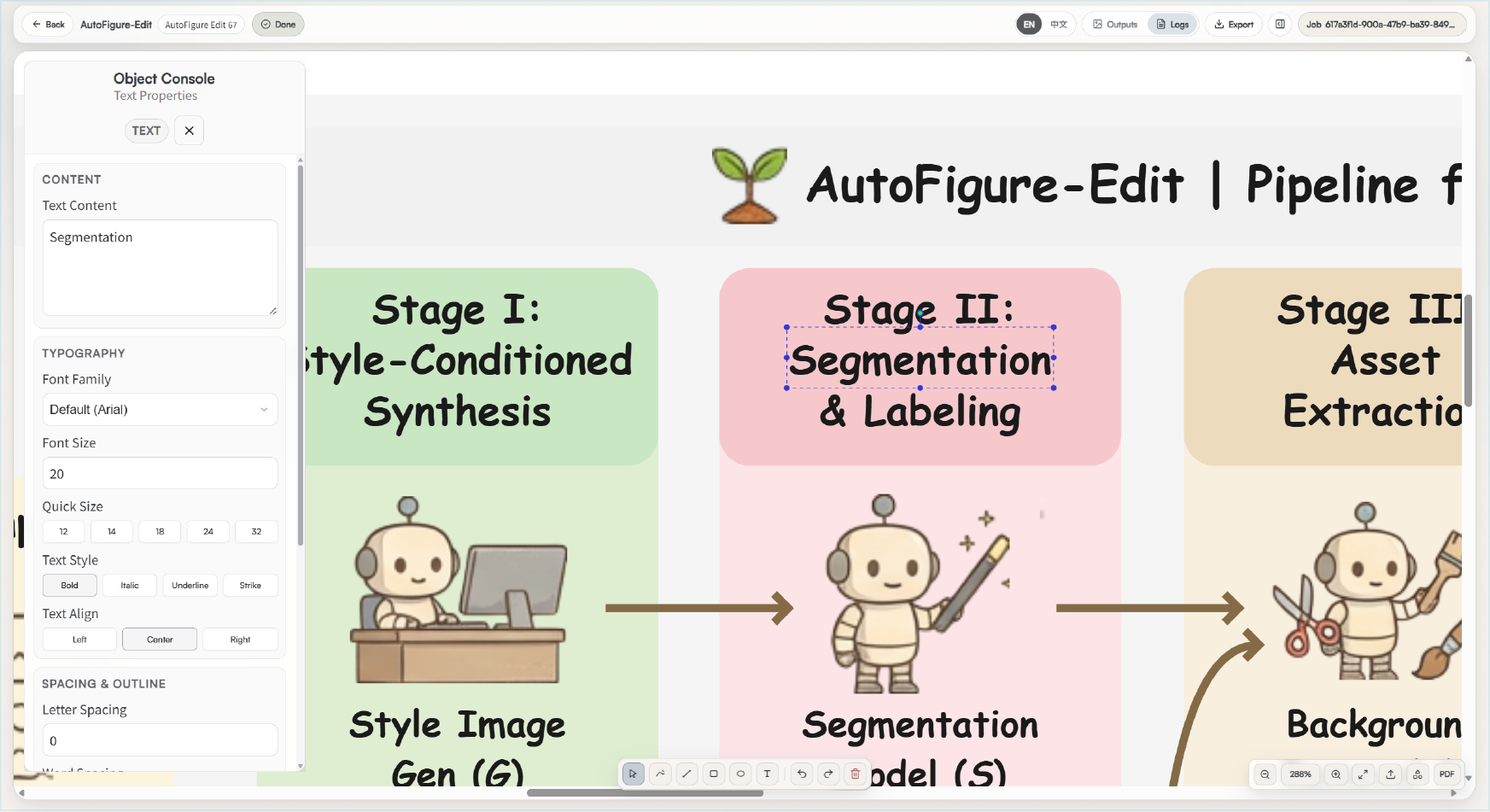}
    \caption{The embedded interactive canvas enables users to freely manipulate individual components within the generated SVG.}
    \label{fig:canvas}
    \vspace{-10pt}
\end{figure}

\textbf{Interactive Editing with SVG Output. }
The generation result of \textsc{AutoFigure-Edit} is a structurally organized SVG file in which semantic elements (e.g., modules, connectors, annotations) are explicitly represented, enabling \textbf{fine-grained manipulation at the component level.} This eliminates the common limitation of raster-based generation, where even minor revisions require regenerating the entire image. More importantly, \textsc{AutoFigure-Edit} further provides \textbf{an embedded Visual Editor} that supports real-time updates. Users can reposition objects, modify text, and adjust stylistic attributes while preserving the overall layout. In Figure~\ref{fig:canvas}, we present the interface of the embedded interactive canvas, which allows users to directly manipulate individual elements within the generated figure.

In summary, \textsc{AutoFigure-Edit} transforms scientific illustration generation into an editable and style-controllable process. For individual researchers, \textsc{AutoFigure-Edit} provides substantial time savings, improved visual clarity, and seamless integration into writing workflows. For the broader community, \textsc{AutoFigure-Edit} promotes more standardized, accessible, and reproducible scientific communication, enabling clearer dissemination of complex ideas.
Additional qualitative results are provided in Figures~\ref{fig:appendix_case_1} and \ref{fig:appendix_case_2}.

\section{Evaluation}
\label{sec:evaluation}
To comprehensively evaluate the usability of \textsc{AutoFigure-Edit}, we conduct (i) automated evaluations on FigureBench~\cite{zhu2026autofigure} and (ii) a user study involving 217 participants.

\subsection{Quantitative Analysis}

\begin{table*}[!t]
\centering
\begin{adjustbox}{width=0.965\textwidth}
\begin{tabular}{@{}lrrrrrrrrr|r@{}}
\toprule
\toprule
& \multicolumn{3}{c}{Visual Design} & \multicolumn{2}{c}{Communication} & \multicolumn{3}{c}{Content Fidelity} & & \\
\cmidrule(lr){2-4} \cmidrule(lr){5-6} \cmidrule(lr){7-9}
\textbf{Method} & Aesthetic & Express. & Polish & Clarity & Flow & Accuracy & Complete. & Appropriate. & \textbf{Overall} & \textbf{Win-Rate} \\
\midrule
\rowcolor[rgb]{ .949,  .949,  .949} 
HTML-Code       & 5.90          & 5.04          & 5.84          & 7.17          & 7.38          & 6.99 & 6.37          & 6.15          & 6.35 & 11.0\% \\
SVG-Code        & 5.00          & 4.19          & 4.89          & 6.34          & 6.48          & 6.15          & 5.53          & 5.37          & 5.49 & 31.0\% \\
\rowcolor[rgb]{ .949,  .949,  .949} 
GPT-Image       & 4.24          & 3.47          & 4.00          & 5.63          & 5.63          & 4.77          & 4.08          & 4.25          & 4.51 & 7.0\% \\
Diagram Agent   & 2.25          & 1.73          & 2.04          & 2.67          & 2.49          & 2.11          & 1.72          & 1.94          & 2.12 & 0.0\% \\
\rowcolor[rgb]{ .949,  .949,  .949}
AutoFigure      & 7.28 & 6.99 & 6.92 & 7.34 & 7.87 & 6.96          & 6.51 & 6.40 & 7.03 & 53.0\% \\
\midrule
\textsc{AutoFigure-Edit}(w/o Ref.)      & \textbf{8.32} & \textbf{8.66} & \textbf{8.16} & 8.10 & \textbf{8.51} & 8.59          & 8.07 & 7.95 & \textbf{8.29} & 76.0\% \\
\rowcolor[rgb]{ .949,  .949,  .949} 
\textsc{AutoFigure-Edit}(with Ref.)      & 7.37 & 7.14 & 7.37 & \textbf{8.15} & 8.43 & \textbf{8.83}          & \textbf{8.26} & \textbf{8.37} & 7.99 & \textbf{83.0\%} \\
\bottomrule
\end{tabular}
\end{adjustbox}
\vspace{-5pt}
\caption{Quantitative comparison of illustration generation methods on FigureBench. Scores are averaged across Visual Design, Communication Effectiveness, and Content Fidelity dimensions. Overall denotes the mean of all sub-metrics, and Win-Rate reflects blind pairwise human preference.}
\label{tab:main_evaluation}
\vspace{-5pt}
\end{table*}

\paragraph{Experimental Setup.}

We adopt the research-paper subset of FigureBench~\cite{zhu2026autofigure} as our evaluation dataset, which contains long-form method sections paired with publication-quality illustrations and provides a realistic testbed for scientific figure generation (Appendix~\ref{sec:appendix_qual}). 

We evaluate on 200 method descriptions sampled from FigureBench: 100 samples are generated without reference style conditioning, and the remaining 100 are generated with reference style images. The style-conditioned subset is further divided into five groups, where each group shares the same reference image, enabling evaluation of style consistency and robustness under fixed stylistic constraints. Example reference styles are provided in Figure~\ref{fig:appendix_case_2}. We compare against three categories of baselines, including end-to-end text-to-image generation (GPT-Image~\citep{hurst2024gpt}), text-to-code generation (HTML-Code and SVG-Code)~\citep{rodriguez2025starvector,malashenko2025leveraging,yang2024matplotagent}, and multi-agent frameworks (Diagram Agent~\citep{wei2025words} and \textsc{AutoFigure}~\cite{zhu2026autofigure}). We use NanoBanana-Pro as the text-to-image model and Gemini-3-Pro as the vision-language model for scaffold synthesis and iterative refinement. The quantitative results are summarized in Table~\ref{tab:main_evaluation}.

\textbf{Overall Performance.}

\textsc{AutoFigure-Edit} consistently outperforms prior approaches across Visual Design, Communication Effectiveness, and Content Fidelity, demonstrating its ability to \textbf{generate publication-quality scientific illustrations and achieving a strong balance} among visual quality and scientific fidelity.

\textbf{Effect of Reference Conditioning.} Reference conditioning reveals a clear trade-off between Visual Design and Content Fidelity.

When reference images are provided, Content Fidelity improves consistently across all three sub-dimensions: Accuracy (8.83), Completeness (8.26), and Appropriateness (8.37), surpassing both the original \textsc{AutoFigure} and the non-conditioned variant and suggesting better semantic grounding for long procedural inputs. In contrast, Visual Design slightly drops (e.g., Aesthetic \textbf{7.37 vs.\ 8.32}), likely because fixed references can constrain stylistic expressiveness.
Despite this, the Win-Rate increases from \textbf{76.0\%} to \textbf{83.0\%}, indicating that reference conditioning yields figures that are more reliably preferred overall. This also suggests that blind pairwise preference is a more holistic and robust indicator than scalar ratings, as it better matches practical selection scenarios.

\subsection{User Study}

\textbf{Experimental Setup.} To assess the real-world usability of \textsc{AutoFigure-Edit}, we conduct a \textbf{deployment-based user study} via our public website\footnote{\url{https://deepscientist.cc/}}.
Users can freely generate editable scientific illustrations and refine the resulting SVGs using the embedded visual editor. Feedback is collected through an integrated interface: once generation completes, a rating dialog is automatically shown, where users evaluate both the rendered figure and its corresponding SVG. All scalar metrics are rated on a 5-point Likert scale (1 = lowest, 5 = highest). We additionally collect a binary \textit{Usability} metric, indicating whether the figure is directly usable in an academic paper without major modifications. Additional details are provided in Appendix~\ref{sec:appendix_eval_details}.

\begin{figure}[t]
    \centering
    \includegraphics[width=\linewidth]{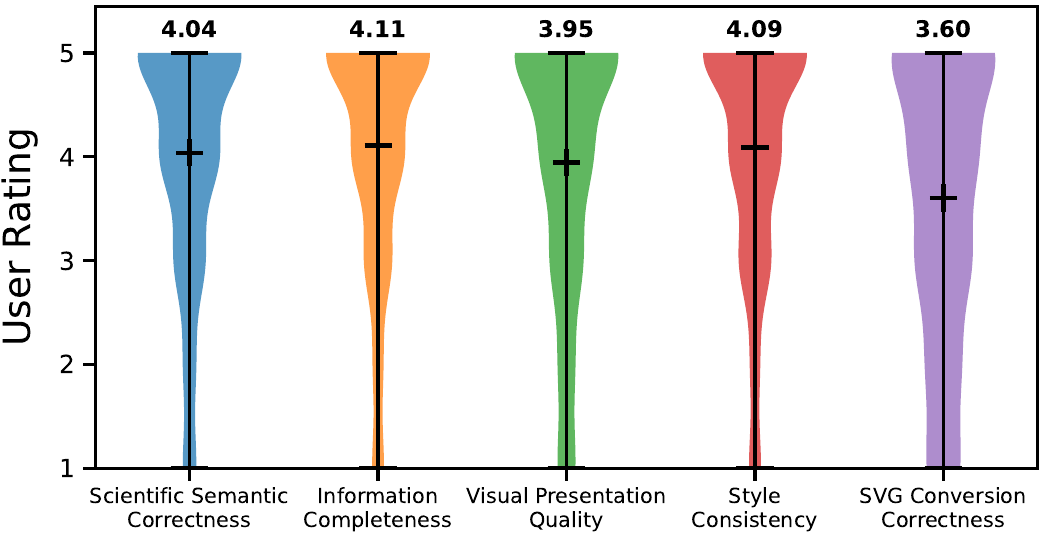}
    \caption{Results of the human user study. The numbers indicate the mean scores. \textsc{AutoFigure-Edit} achieves consistently high satisfaction in most metrics.}
    \label{fig:user_study}
    \vspace{-15pt}
\end{figure}

\textbf{Result Analysis.} We collect 262 evaluation samples from 217 unique participants. The aggregated results and rating distributions are illustrated in Figure~\ref{fig:user_study}. For the generated PNG figures, \textbf{\textsc{AutoFigure-Edit} achieves strong performance across all evaluation dimensions}, with mean scores of 4.04 (Scientific Semantic Correctness), 4.11 (Information Completeness), 3.95 (Visual Presentation Quality), and 4.09 (Style Consistency). Notably, \textbf{ratings are heavily concentrated at the highest level}: 48\% and 51\% of evaluations assign Score 5 to semantic correctness and completeness, respectively, and 50\% to style consistency. Low ratings (Score 1–2) are rare for semantic dimensions (generally below 12\%), demonstrating that 
the system reliably preserves scientific meaning and structural integrity across diverse user inputs.

\textbf{Practical usability} further confirms the system’s effectiveness. \textbf{126} out of 262 users consider the generated figure directly suitable for inclusion in an academic paper without additional modification. Given that direct usability requires conceptual correctness, satisfactory layout and stylistic quality, this result \textbf{highlights the system’s readiness for real-world research workflows} rather than merely benchmark-level adequacy.

For PNG-to-SVG reconstruction, the average Conversion Correctness score reaches 3.60, with the majority of evaluations concentrated in the upper-middle to high range (Scores 3–5) and 36\% achieving Score 5. Very low scores remain limited, indicating that catastrophic structural failures are uncommon. While minor geometric deviations may occasionally arise during reconstruction, the fully editable SVG output ensures that such issues can be corrected with minimal manual effort, thereby preserving downstream usability.

Overall, the empirical results show that \textsc{AutoFigure-Edit} achieves strong semantic fidelity, high informational completeness, and substantial real-world usability in deployment, suggesting it can be effectively integrated into academic figure production workflows.

\section{Conclusion}
In this paper, we presented \textsc{AutoFigure-Edit},an end-to-end system that generates fully editable scientific illustrations from long-form text with reference-guided style adaptation and native SVG editing. Quantitative evaluations and a deployment-based user study showed that \textsc{AutoFigure-Edit} consistently outperformed prior methods and produced outputs that users frequently judged ready for acadamic publication.

\section*{Limitations}

While our deployment and user study highlight the practical utility of \textsc{AutoFigure-Edit} in scientific illustration generation, some limitations remain:

\textbf{Dependence on foundation models.} Our pipeline currently relies on closed-source vision and vision-language models (e.g., Gemini-3.1-Pro-Preview and NanoBanana-Pro) for style-conditioned raster synthesis and SVG template refinement. This reliance may incur usage costs, raise data privacy concerns, and limit the reproducibility of our pipeline. While current open-source alternatives struggle with the complex spatial reasoning required for this specific task, future work will explore integrating powerful open-weight models to mitigate these accessibility limitations once their capabilities sufficiently mature.

\textbf{Error propagation.} As \textsc{AutoFigure-Edit} derives vector structures from an intermediate raster draft, upstream segmentation errors (e.g., incorrectly merged split visual components) can cascade through the pipeline,
requiring manual adjustments via the embedded editor.

\textbf{Scope and evaluation constraints.} The embedded visual editor is designed for localized, component-level refinements and is not intended to replace comprehensive graphic design software. Furthermore, our current user study was primarily intended as a usability evaluation to assess real-world workflow efficiency. Validating the system across a wider variety of highly specialized scientific domains and conducting rigorous expert-only correctness checks remains an area for future work.

We hope that our system will set a new standard for automated figure generation workflows, bridging the gap between complex scientific concepts and accessible, high-quality visual communication.

\section*{Ethics and Broader Impact Statement}

We acknowledge the significant ethical considerations associated with powerful generative technologies like \textsc{AutoFigure-Edit}. The primary risk involves the potential for misuse, where the system could be used to generate scientifically plausible but factually incorrect or misleading schematics to support false claims. To mitigate this risk, we are committed to transparency and responsible deployment. Our mitigation strategy is twofold. First, we explicitly declare that \textsc{AutoFigure-Edit} is an assistive tool with limitations. This disclaimer, stating that the system is not a substitute for expert verification and may not produce perfectly reliable outputs, will be prominently placed in this paper and in the README file of the public code repository. Second, the open-source license governing \textsc{AutoFigure-Edit} will include a mandatory attribution clause. This clause requires any academic publication using a figure generated by our tool to (a) include a specific section that discusses the role AI played in the work, and (b) explicitly caption the figure as having been generated by \textsc{AutoFigure-Edit}. These requirements are designed to ensure transparency and accountability in downstream use of our technology, fostering a research environment where AI tools augment, rather than compromise, scientific integrity.

\bibliography{custom}

\clearpage

\appendix

\section{Quantitative Evaluation}
\label{sec:appendix_qual}

\begin{figure*}[h]
    \centering
    \includegraphics[width=0.92\linewidth]{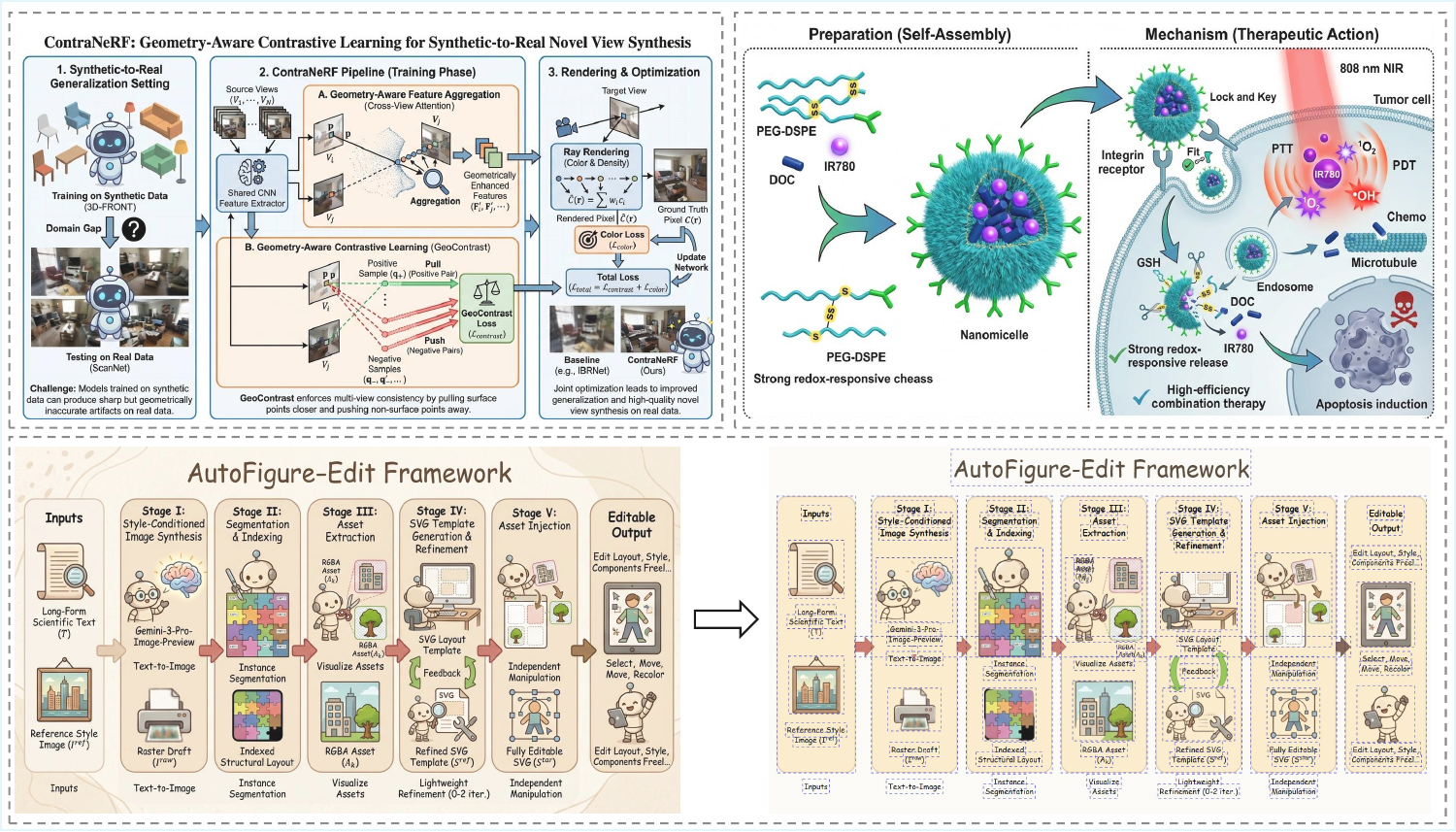}
    \caption{Qualitative results of \textsc{AutoFigure-Edit}.}
    \label{fig:appendix_case_1}
\end{figure*}

\begin{figure*}[htbp]
    \centering
    \includegraphics[width=0.92\linewidth]{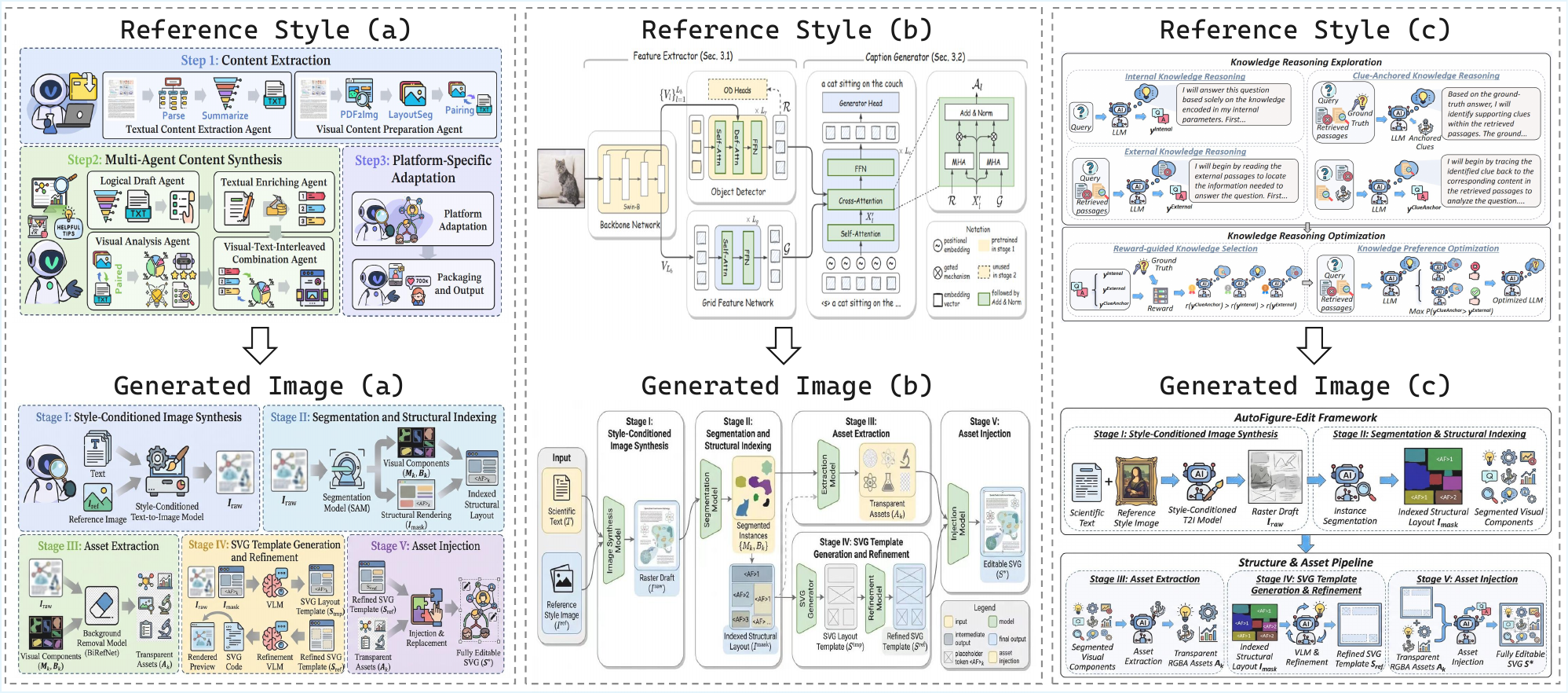}
    \caption{Comparison between Reference Images and Generated Images.}
    \label{fig:appendix_case_2}
\end{figure*}

We compare against three categories of baselines: (1) \textbf{end-to-end text-to-image} methods~\citep{sun2024autoregressive}, where we use GPT-Image~\citep{hurst2024gpt} to directly generate a scientific schematic from the paper text following standardized instructions; (2) \textbf{text-to-code} methods, where an LLM produces HTML and SVG code~\citep{rodriguez2025starvector,malashenko2025leveraging,yang2024matplotagent} that is automatically rendered into images; and (3) \textbf{multi-agent} frameworks, represented by Diagram Agent~\citep{wei2025words} and \textsc{AutoFigure}~\cite{zhu2026autofigure}. In our experiments, we employ FigureBench as evaluation benchmark, which adopts a VLM-as-a-judge paradigm designed for structural reasoning and long-context scientific illustration assessment, and reports referenced, multi-dimensional scores. Evaluation follows a multi-dimensional rubric covering \textbf{Visual Design} (aesthetic quality, visual expressiveness, professional polish), \textbf{Communication Effectiveness} (clarity, logical flow), and \textbf{Content Fidelity} (accuracy, completeness, appropriateness). We additionally report \textbf{Win-Rate}, computed via blind pairwise comparisons against the reference illustrations, measuring how often a method is preferred as producing the more suitable figure for a given description.

\section{User Study}

\label{sec:appendix_eval_details}

The user study consists of two complementary parts: (i) Figure Evaluation, which assesses the quality of the originally generated figure, and (ii) SVG Conversion Evaluation, which evaluates the fidelity of the converted SVG.

All scalar metrics are rated on a 5-point Likert scale (1 = lowest, 5 = highest). The process of evaluation is conducted through an integrated feedback interface on our website. After \textsc{AutoFigure-Edit} generates the final output, a rating dialog is automatically presented to the user. If the user dismisses the dialog, it will reappear when the user attempts to download the generated figure; submission of the evaluation is required prior to download. This design ensures that all collected ratings correspond to actual usage scenarios and are provided by users who have interacted with the generated result. 
To minimize ambiguity and promote consistent interpretation of the evaluation criteria, each rating dimension in the feedback dialog is accompanied by a small “?” icon. Clicking this icon opens a detailed scoring guideline that standardizes the evaluation process. The guideline is structured into three components: a \textbf{Definition} that formalizes the metric, a \textbf{Guiding Question} that anchors user judgment, and a \textbf{Scoring Rubric} that specifies the semantic meaning of each score level. This design helps align user understanding with the intended evaluation protocol and reduces variance caused by subjective interpretation. The detailed scoring guidelines are provided as follows:

\subsection*{Part I: Figure Evaluation (PNG)}

\textbf{Scientific Semantic Correctness (1–5).}  
(i) \textbf{Definition}: Measures whether the figure accurately represents the scientific concepts, processes, and
relationships described in the input method text. 
(ii) \textbf{Question}: Does the figure correctly reflect the scientific content described in the method section?
A score of 5 indicates fully correct semantic representation, while 1 indicates a misleading or incorrect depiction.

\textbf{Information Completeness (1–5).}  
(i) \textbf{Definition}: Measures whether all key components and steps described in the method text are present in the
figure. 
(ii) \textbf{Question}: Are all essential elements from the method text included in the figure?
Higher scores indicate more comprehensive coverage of essential elements.

\textbf{Visual Presentation Quality (1–5).}  
(i) \textbf{Definition}: Evaluates the visual clarity, readability, and overall professionalism of the figure. 
(ii) \textbf{Question}: Is the figure visually clear, well-aligned, and suitable for academic publication?
Users should consider whether the figure would be suitable for academic publication.

\textbf{Style Consistency (1–5).}  
(i) \textbf{Definition}: Measures how well the generated figure matches the style of the provided reference image.
(ii) \textbf{Question}: Does the generated figure follow the visual style of the reference figure? 

\textbf{Usability (Binary: 0/1).}  
(i) \textbf{Definition}: Determines whether the figure is directly usable in an academic paper without major modifications.
(ii) \textbf{Question}: Can this figure be directly used in a paper?

\subsection*{Part II: SVG Conversion Evaluation}

\textbf{Conversion Correctness (1–5).}  
(i) \textbf{Definition}: Measures whether the structural and semantic elements are correctly preserved during the png-to-SVG conversion. This includes: (a) Correct placement of components; (b) Correct correspondence between original objects and SVG elements; (c) No missing or duplicated elements
(ii) \textbf{Question}: Are all elements correctly preserved and positioned after conversion to SVG?

\section{License}

\textsc{AutoFigure-Edit} is released under the MIT License. Users are solely responsible for any content generated using our demo (including the resulting figures and SVGs), as well as for ensuring compliance with applicable laws, third-party rights, and publication policies. We do not assume liability for any direct or indirect damages arising from the use of the software or from the generated outputs.

\end{document}